\documentclass[11pt]{article}
\usepackage{url}    

\usepackage{graphicx}
\usepackage{float}
\usepackage{rotating}
\usepackage{tabularx}

\usepackage[british]{babel}		

\usepackage{amssymb}					
\usepackage{amsfonts}					
\usepackage{amsmath}					
\usepackage[mathscr]{euscript}

\let\oldmarginpar\marginpar

\renewcommand\marginpar[1]{\-\oldmarginpar[\raggedleft\small\textsf #1]{\raggedright\small\textsf #1}}

\newcommand{\XX}{\ensuremath{\mathbf{X}}}

\newcommand{\yy}{\ensuremath{\mathbf{y}}}

\title{Learning under Concept Drift: an Overview}

\author{Indr\.e \v{Z}liobait\.e\\
Faculty of Mathematics and Informatics\\
Vilnius University, Lithuania
\texttt{zliobaite@gmail.com}\\
}

\date{}

\begin{document}
\maketitle 

Concept drift refers to a non stationary learning problem over time. The training and the application data often mismatch in real life problems \cite{Hand06}. 

In this report we present a context of concept drift problem \footnote{This is a working version, the categorization is in progress. The latest version of the report is available online: \url{http://zliobaite.googlepages.com/Zliobaite_CDoverview.pdf} . Feedback is very welcome.}. 
We focus on the issues relevant to adaptive training set formation.  
We present the framework and terminology, and formulate a global picture of concept drift learners design. 

We start with formalizing the framework for the concept drifting data in Section \ref{sec:framework}. 
In Section \ref{sec:designassumptions} we discuss the adaptivity mechanisms of the concept drift learners. 
In Section \ref{sec:taxonomy} we overview the principle mechanisms of concept drift learners. 
In this chapter we give a general picture of the available algorithms and categorize them based on their properties. 
Section \ref{sec:applications} discusses the related research fields and Section \ref{sec:applications} groups and presents major concept drift applications.

This report is intended to give a bird's view of concept drift research field, provide a context of the research and position it within broad spectrum of research fields and applications. 

\section{Framework and Terminology}
\index{terminology}
\label{sec:framework}

For analyzing the problem of training set formation under concept drift, we adopt the following framework. 

A sequence of instances is observed, one instance at a time, not necessarily in equally spaced time intervals. Let $\XX_t\in{\Re}^{p}$ is a vector in $p$-dimensional feature space observed at time $t$ and $\yy_t$ is the corresponding label. For classification $\yy_t\in{\cal Z}^{1}$, for prediction $\yy_t\in{\Re}^{1}$. We call $\XX_t$ \emph {an instance} and a pair $(\XX_t,\yy_t)$ \emph{a labeled instance}. 
We refer to instances $(\XX_1,\ldots,\XX_t)$ as \emph{historical data} and instance $\XX_{t+1}$ as \emph{target} (or testing) instance. 

\subsection{Incremental Learning with Concept Drift}

\begin{figure}
\centering
\includegraphics[width=0.8\textwidth]{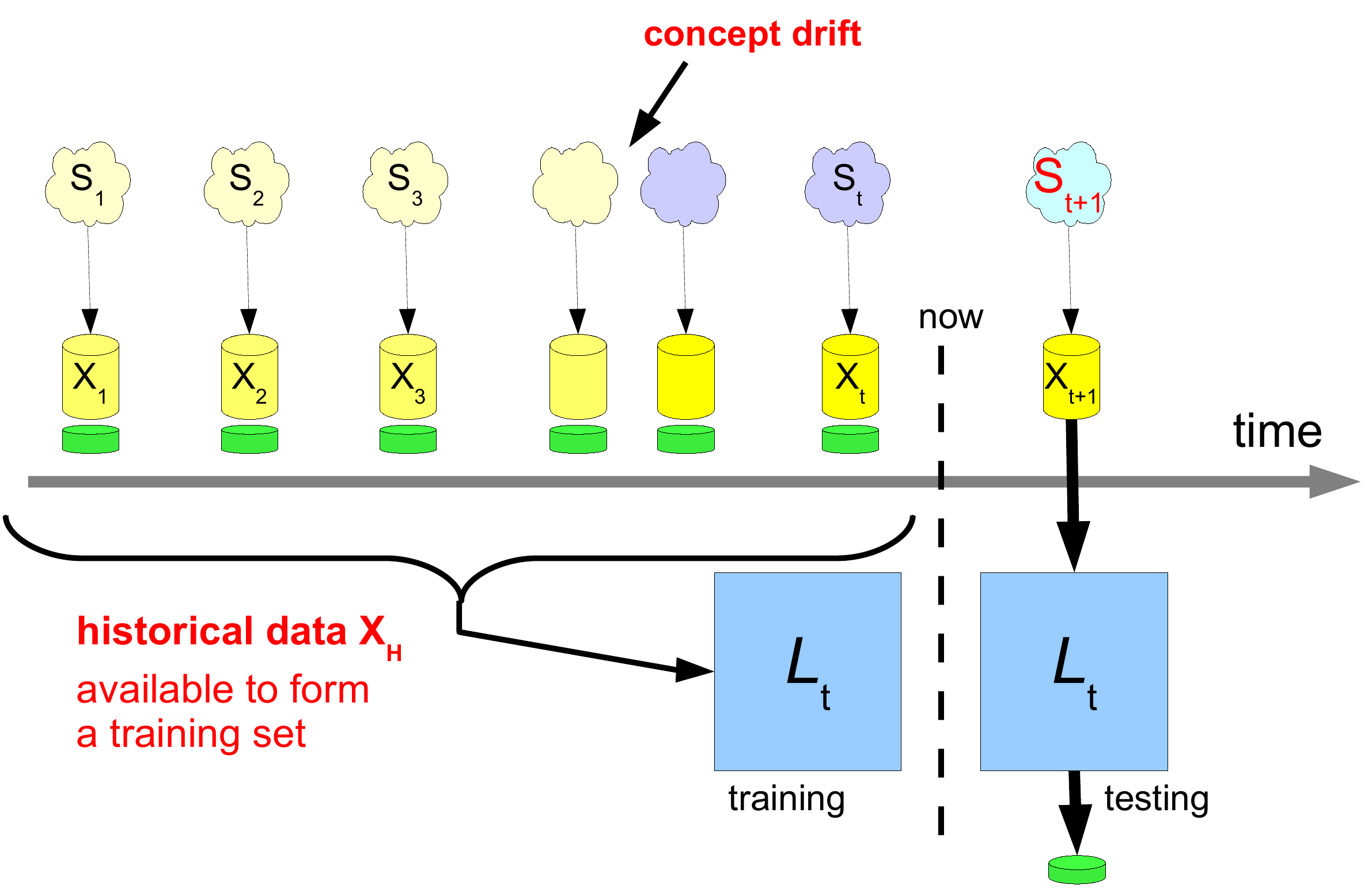}
\caption{One time step (t) of the incremental learning process}
\label{fig:framework}	
\end{figure}

We use incremental learning framework. At every time step $t$ we have historical data (labeled) 
available ${\bf X^H}=(\XX_1,\ldots,\XX_t)$. 
A target instance $\XX_{t+1}$ arrives. The task is to predict a label $\yy_{t+1}$. For that we build a learner ${\cal L}_t$, using all or \emph{a selection} from the available historical data ${\bf X^H}$. We apply the learner ${\cal L}_t$ to predict the label for $\XX_{t+1}$. A prediction process at time step $t$ is illustrated in Figure \ref{fig:framework}. That is for one time step. 

\begin{figure}[t]
\centering
\includegraphics[width=0.9\linewidth]{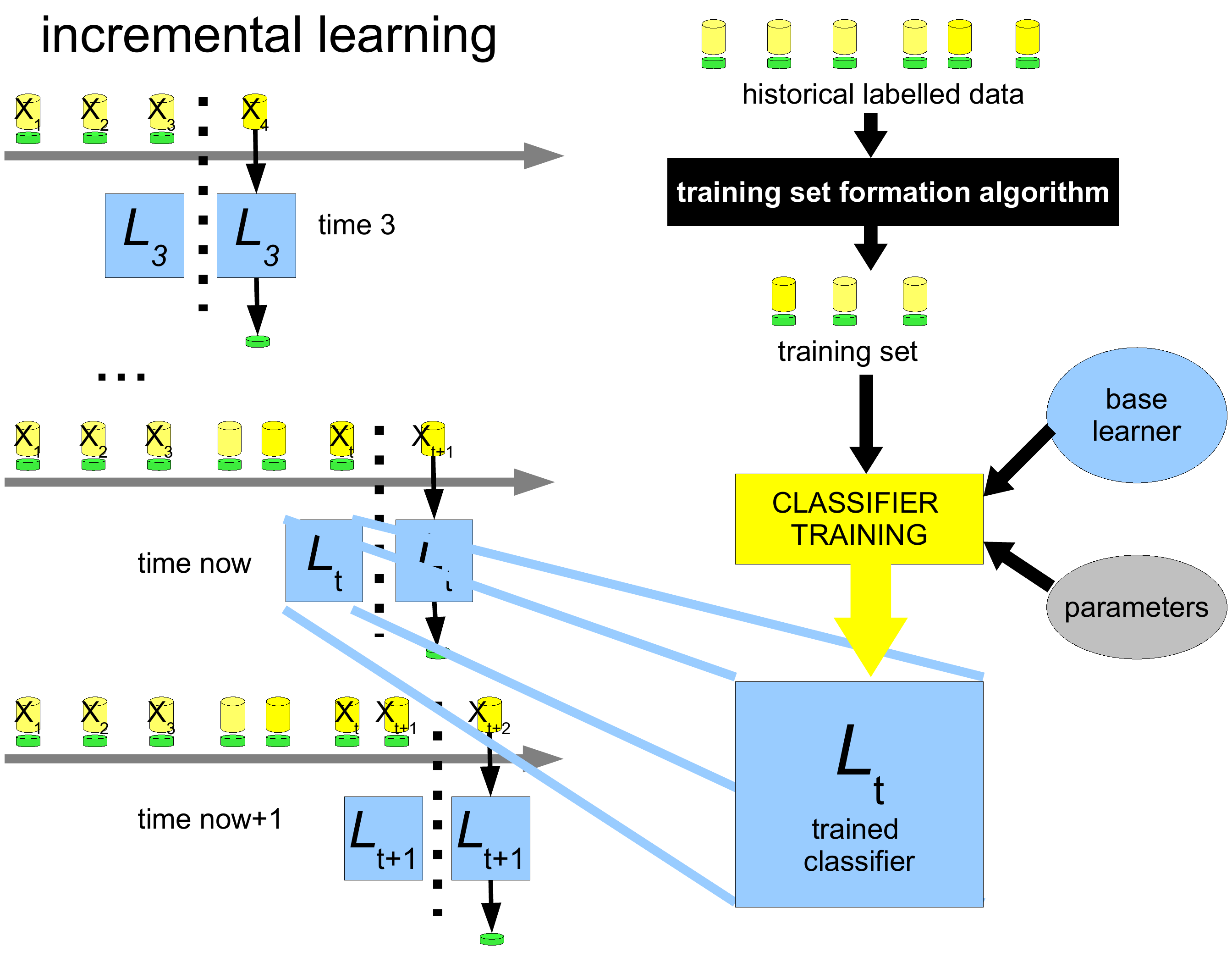}
\caption{Incremental learning process}
\label{fig:incrementallearning}	
\end{figure}

At the next step after the classification or prediction decision is casted, the label $\yy_{t+1}$ becomes available. How the instance $\XX_{t+1}$ with a label is a part of historical data. 
The next testing instance $\XX_{t+2}$ is observed. 
We picture a fragment of the incremental learning loop in Figure \ref{fig:incrementallearning}.
The classifier training phase at time $t$ is zoomed in. 
Training set formation strategies are the subjects of our investigation. 
They are depicted as a `black box' in the figure.

Every instance $\XX_t$ is generated by a source $S_t$. We delay more formal definition of a source until the next section, for now assume that it is a distribution over the data. If all the data is sampled from the same source, i.e. $S_1 = S_2 = \ldots = S_{t+1}= {\bf S}$ we say that the \emph{concept} is stable. If for any two time points $i$ and $j$ $S_i\neq S_j$, we say that there is a \emph{concept drift}. 
\index{concept drift}

Note that a random noise (deviation) is not considered to be a concept drift, because the data generating source is still the same. 

\textbf{The core assumption when dealing with the concept drift problem is uncertainty about the future.} 
We assume that the source of the target instance $\XX_{t+1}$ is not known with certainty. 
It can be assumed, estimated or predicted but {\bf there is no certainty}. 
Otherwise the data can be decomposed into two separate data sets and learned as individual models or in a combined manner (then it is a multitask learning problem \cite{Ben-David08}).

We do not consider \emph{periodic} seasonality as concept drift problem. But if seasonality is \emph{not known} with certainty, we consider it as concept drift problem. For instance, a peak in sales of ice cream is associated with summer but it can start at different time every year depending on the temperature and other factors, therefore it is not known exactly when the peak will start.

\subsection{Causes of a concept drift}
\index{causes of a concept drift}
\label{sec:changesource}

Before looking what can actually cause the drift, let us return to the source $S_t$ and provide more rigorous definition of it. 


Classification problem independently of presence or absence of concept drift may be described as follows \cite{Narasimurty07}. 
Let $\XX\in{\Re}^{p}$ is an instance in $p$-dimensional feature space. 
$\XX\in c_i$, where ${c_1, c_2,\ldots ,c_k}$ is the set of class labels. 
The optimal classifier to classify $X\rightarrow c_i$ is completely determined by a prior probabilities for the classes $P(c_i)$ and the class-conditional probability density functions (pdf) $p(\XX|c_i)$,  $i=1,\ldots,k$. 

We define a set of a prior probabilities of the classes and class-conditional pdf's as concept or \emph{data source}:
\begin{equation}
	{\bf S}=\{(P(c_1),p(\XX|c_1)),(P(c_2),p(\XX|c_2)),\ldots,(P(c_k),p(\XX|c_k))\}.
\end{equation}
When referring to a particular source at time $t$ we will use the term \emph{source}, while when referring to a fixed set of prior probability and the classes and class-conditional pdf we will use the term \emph{concept} and denote it ${\bf S}$.

Recall, that in Bayesian decision theory \cite{Duda} the classification decision for instance $\XX$ at equal costs of mistake is made based on maximal a posteriori probability, which for a class $c_i$ is:
\begin{equation}
p(c_i|\XX) = \frac{P(c_i)p(\XX|c_i)}{p(\XX)},
\label{eq:bayesianclassification}
\end{equation}
where p(\XX) is an evidence of $\XX$, which is constant for all the classes $c_i$.

As first presented by Kelly et al \cite{Kelly99}, concept drift may occur in thee ways. 
\begin{enumerate}
\item Class priors $P(c)$ might change over time.
\item The distributions of one or several classes $p(\XX|c)$ might change.
\item The posterior distributions of the class memberships $p(c|\XX)$ might change.
\end{enumerate}
Note, that the distributions $p(\XX|c)$ might change in such a way that the class membership is not affected (e.g. symmetric movement to opposite directions). 

Sometimes change in $p(\XX|c)$ (independently whether it affects $p(c|\XX)$ or not) is referred as virtual drift and change in $p(c|\XX)$ is referred as real drift \cite{Widmer93}. We argue, that from practical point of view it is not essential whether the drift is real or virtual, since $p(c|\XX)$ depends on $p(\XX|c)$ as in Equation (\ref{eq:bayesianclassification}). In this thesis now on we do not make a distinction between the real and virtual drifts.


\section{How Do Concept Drift Learners Work?}
\label{sec:designassumptions}

Following the framework, which was set-up in the previous section, the learner should provide the most accurate generalization for the data at time $t+1$. In order to build such a learner, four main design sub-problems need to be solved. 

\begin{description}
\item[A.1 Future assumption:] a designer needs to make an assumption about the future data source $S_{t+1}$.
\item[A.2 Change type:] a designer needs to identify possible change patterns.
\item[A.3 Learner adaptivity:] based on the change type and the future assumption a designer chooses the mechanisms which make the learner adaptive. 
\item[A.4 Model selection:] a designer needs a criterion to choose a particular parametrization of the selected learner at every time step (e.g.\@ the weights for an ensemble members, the window size for variable window method).
\end{description}

\begin{figure}[t]
\centering
\includegraphics[width=0.8\linewidth]{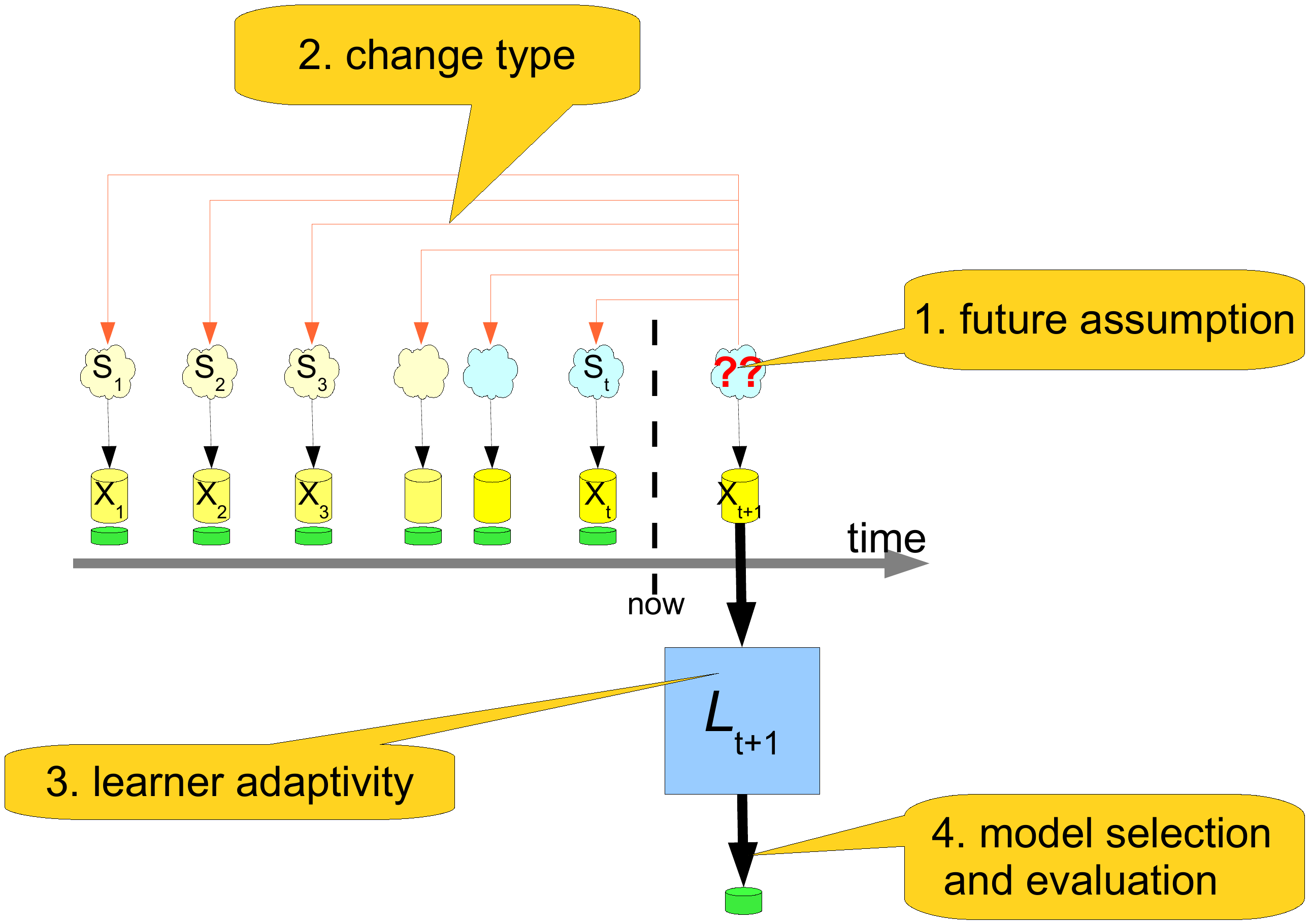}
\caption{Sub-problems of the concept drift learner design.}
\label{fig:positioning}	
\end{figure}

All these sub-problems are the choices to be made when designing a learner. 
In Figure \ref{fig:positioning} we depict a positioning of each design sub-problem within the established learning framework.

In the next subsections we discuss each of the design sub-problems individually.

\subsection{Future assumption}

Future assumption is the assumption to be made about the source $S_{t+1}$ of the target instance $\XX_{t+1}$. We identify three types of choices here. 
\begin{enumerate}
\item Assuming that $S_{t+1} = S_t$. 
\item Estimating the source based on $\XX_{t+1}$.
\item Predicting the change.
\end{enumerate}

The first option, assuming $S_{t+1} = S_t$, is the most common among concept drift, although rarely explicitly stated. It is assumed that in the nearest future we will see the data coming from the same source as we saw in the near past. 

The second option utilizes information from the unlabeled target instance $\XX_{t+1}$. Estimation of the source is usually done by measuring the distance between $\XX_{t+1}$ and historical reference instances. The algorithms presented in \cite{Tsymbal08,Pourkashani08,Delany05,Lazarescu03} use this future assumption. 

Generally in concept drift problem the future data source is not known with certainty. However, there are methods using trainable prediction rules, to estimate the future state and incorporate that estimation into the incremental learning process. The algorithms using future predictions are presented in \cite{Brown09,Wu05,Yang06,Bottcher08}.

\subsection{Change types}

In Section \ref{sec:changesource} we identified the causes of a drift, or what happens to the data generating source itself. 
Here by change types we mean the configuration patterns of the data sources over time. 
The structural types of change are usually defined based on those configurations. 

For intuitive explanation, let us for now restrict the number of possible sources over time to two: $S_I$ and $S_{II}$.

The simplest pattern of a change is \emph{sudden drift}, when at time $t_0$ a source $S_I$ is suddenly replaced by source $S_{II}$. For example, Kate is reading the news. Sudden interest in meat prices in New Zealand when she got an assignment to write an article, is a sudden drift.

\emph{Gradual drift} is another type often met in the literature. However in fact there are two types being mixed under this term. The first type of gradual drift is referring to a period when both sources $S_I$ and $S_{II}$ are active (e.g. \cite{Stanley03,Widmer96,Narasimurty07}). As time passes, the probability of sampling from source $S_I$ decreases, probability of sampling from source $S_{II}$ increases. Note, that at the beginning of this gradual drift, before more instances are seen, an instance from the source $S_{II}$ might be easily mixed up with random noise.

Another type of drift also referred as gradual includes more than two sources, however, the difference between the sources is very small, thus the drift is noticed only when looking at a longer time period (e.g. \cite{Tsymbal08,Delany05,Garcia06}). 
We refer to the former type of gradual drift as \emph{gradual} and the latter type of drift as \emph{incremental} (or stepwise).
For example, gradual drift is increasing interest in real estate, while Kate prefers real estate news more and more over time when her interest in buying a flat increases.

Finally, there is another big type of drift referred as \emph{reoccurring context}. That is when previously active concept reappears after some time. It differs from common seasonality notion in a way that it is not certainly periodic, it is not clear when the source might reappear. In Kate's example these are the biographies of Formula-1 drivers. The interest is related to the schedule of the races. But she does not look up the biographies at the time of the races, because she is watching them at the time. She might want to look up them later in the middle of the week. And the particular drivers she will be interested in might depend on who won the races this time.

\begin{figure}[t]
\centering
\includegraphics[width=0.7\linewidth]{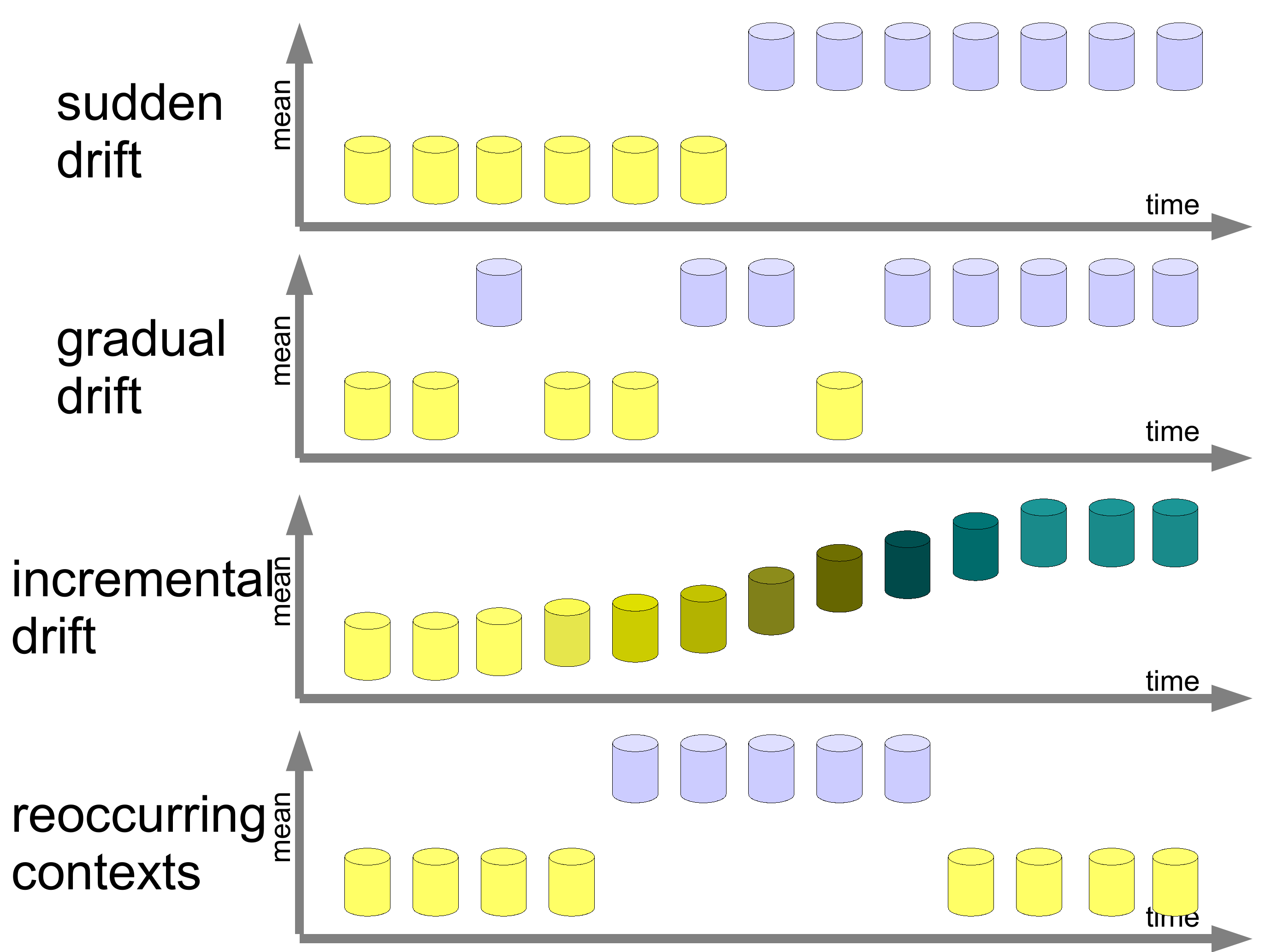}
\caption{Illustration of the four structural types of the drift.}
\label{fig:drifts}	
\end{figure}

In Figure \ref{fig:drifts} we give an illustration of the main structural drift types, assuming one dimensional data, where a source is characterized by the mean of the data. We depict only the data from one class.

Note that the types of drifts discussed here are not exhaustive. 
If we think of a data segment of length $t$ and just two data generating sources $S_I$ and $S_{II}$, the number of possible combinations of the sources (that means possible change patterns) would be $2^{t}$, a lot.
Moreover, in concept drift research it is often assumed that the data stream is endless, thus there could be infinite number of possible change patterns. We define the major structural types, since we argue, that assumption about the change types is absolutely needed for designing adaptivity strategies. 

Recently there has been an attempt to categorize change types into mutually exclusive categories \cite{Minku09} based on number of reoccurences, severity, speed and predictability. 
In principle the proposed categorization tires to quantify the main aspects of the learner design process into change categorization. 
We argue that the categories cannot be mutually exclusive, because the change frequency count, speed, severity is relative to the length of the subsequence, at which one is looking. 
Thus we restrict our categorization to very few qualitative categories. 
We reserve predictability as a part of the learner design process (future assumption), not the change itself. 

\subsection{Learner adaptivity}
\label{sec:2:adaptivity}

We identify four main adaptivity areas:
\begin{enumerate}
\item Base learners can be adaptive (e.g. configuration of decision tree nodes \cite{Hulten01}).
\item Parametrization of the learners can be adaptive (e.g. weighting training samples in support vector machines \cite{Klinkenberg00}).
\item Adaptive training set formation (e.g. training windows, instance selection) can be employed, which is the scope and focus of this thesis. Training set formation can be decomposed into
	\begin{itemize}
	\item training set selection,
	\item training set manipulation (e.g. bootstrapping, noise),
	\item feature set manipulation.
	\end{itemize}	
\item Fusion rules of the ensembles (\cite{Street01,Wang03,Stanley03}).
\end{enumerate}

The adaptivity strategies, which are based on training set selection selection, can be generally divided into windowing (selecting training instances consecutive in time) and instance selection (when sequential in time instances are selected as a training set). The choice of adaptivity strategy strongly depends on the assumption about the change type, discussed in the previous section. For sudden drift windowing strategies are generally preferred, while for gradual drift and reoccurring contexts instance selection strategies are preferred. 

\subsection{Model selection}

In this thesis we use the generalization error as the primary measure of the concept drift learner performance. Thus for model selection (training) purposes the procedure of estimation the expected generalization error for target instance $\XX_{t+1}$ at every time step needs to be defined. The two main options are:
\begin{enumerate}
\item theoretical evaluation of the generalization error, and
\item estimation of the generalization error using cross validation.
\end{enumerate}
In any case error estimation choice is strongly related to the future assumption, because it depends on the expectation regarding the future data source $S_{t+1}$.

\begin{figure}[t]
\centering
\includegraphics[width=0.7\linewidth]{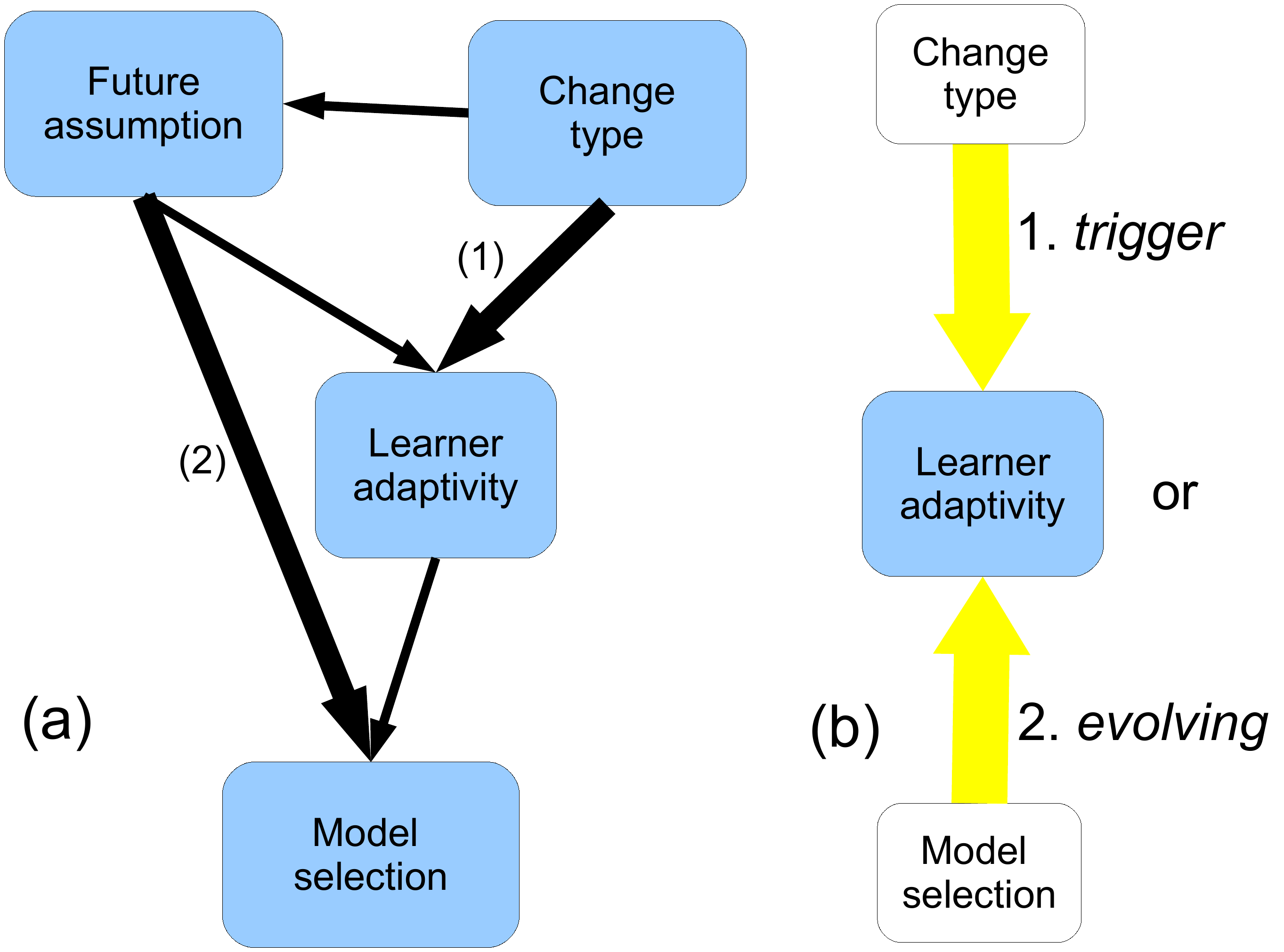}
\caption{The design process of a concept drift learner.}
\label{fig:design}	
\end{figure}

The design process of a concept drift learner is graphically illustrated in Figure \ref{fig:design} (a). We see relations (1) and (2) as key issues in designing concept drift learners. (1) the strategies selected to make the learners adaptive would strongly depend on the assumption about the change type, present in the data. (2) the model selection and evaluation strategies would strongly depend on the assumption about the future data source, on which the learner will be applied.

It is common to categorize concept drift learners into two major groups: 
\begin{enumerate}
\item learner adaptivity is initiated by a \emph{trigger} (or active change detector), and 
\item a learner regularly \emph{evolves} independently of the alarms or detectors. 
\end{enumerate}
The two categories can be positioned within the design framework we just defined. In the first group the initiation for learner adaptivity comes from the `change type' block, while in the second group the adaptivity is based on `model evaluation and selection' block. The process is illustrated in Figure \ref{fig:design} (b).

We will give more details about the categories of the drift learners in the next section, where we overview the related work.

\section{Taxonomy of Available Concept Drift Learners}
\label{sec:taxonomy}
\index{concept drift learners, taxonomy}

In this section we overview and map the related work. This section is intended to give a general view, the approaches specifically related to our work will be presented in corresponding chapters. The overview is concentrated on a supervised learning under concept drift. 

Schlimmer and Granger \cite{Schlimmer86} in 1986 formulated the problem of incremental learning from noisy data and presented an adaptive learning algorithm STAGGER. They are the authors of the term `concept drift'. 
Since then a number of studies dealing with concept drift problem appeared. There were three `peaks' in interest, one around 1998 followed by a special issue of Machine Learning journal \cite{ML98}, the other around 2004 followed by a special issue of Intelligent Data Analysis journal \cite{IDA04}. The third `peak' started around 2007 and continues now on, as a result of increasing loads of streaming data and computational resources.  Several PhD theses have directly addressed the problem of concept drift \cite{Bifetthesis,Nishidathesis,Widyantorothesis,Castillothesis, Spinosathesis}. 

The learners responsive to a concept drift can be divided into two big groups based on \emph{when} the adaptivity is `switched on'. They are either trigger based or evolving. Trigger based means that there is a signal which indicates a need for model change. The trigger directly influences \emph{how the new model should be constructed}. Most often change detectors are employed as triggers. 
The evolving methods on the contrary \emph{do not maintain an explicit link between the data progress and model construction} and usually do not detect changes. 
They aim to build the most accurate classifier either by maintaining the ensemble weights or prototyping mechanisms. They usually keep a set of alternative models, and the models for a particular time point are selected based on their performance estimation. 
This is `why' dimension in the taxonomy.

Another dimension for grouping concept drift learners is based on \emph{how} the learners adapt. 
What are the actual adaptation mechanisms? 
The mechanisms were discussed following the design assumption A.4 presented in Section \ref{sec:designassumptions}. Generally the adaptation mechanisms are either related to training set formation or a design and parametrization of the base learner. 

\begin{figure}[t]
\centering
\includegraphics[width=0.8\linewidth]{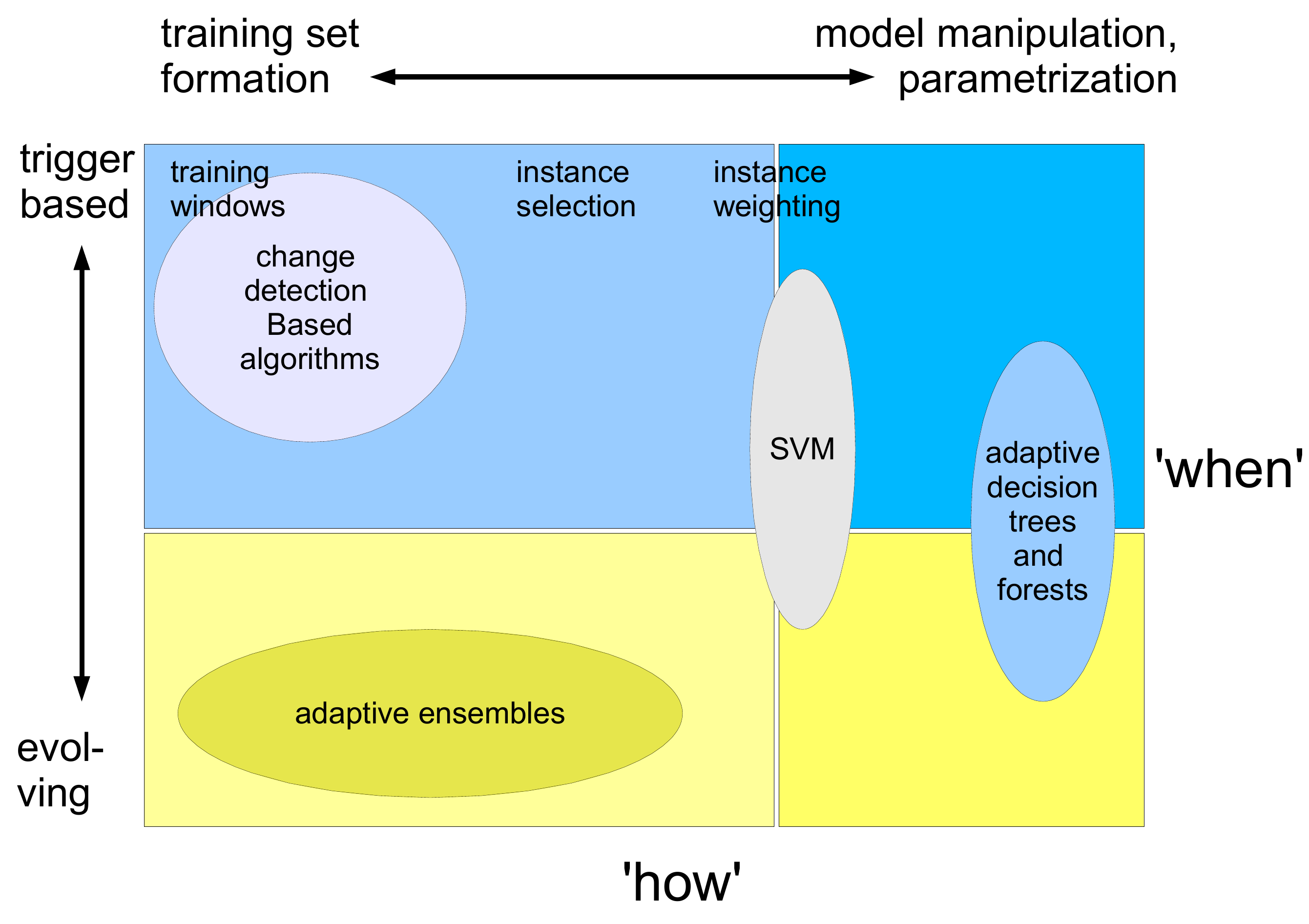}
\caption{A taxonomy of adaptive supervised learning techniques.}
\label{fig:taxonomy}	
\end{figure}

Based on those two dimensions we overview the main methodological contributions available in the literature. 
The taxonomy is graphically presented in Figure \ref{fig:taxonomy}. 
The positions of popular techniques (our interpretation) are indicated by ellipses.

\subsection{Evolving learners}


We start by overviewing the evolving techniques. Some of the techniques discussed above employ change detection mechanisms, still these are not the triggers of adaptation (`detect and cut'), but rather a tool to reduce computational complexity. First we discuss ensemble techniques, which make the largest group, and then other evolving techniques. 

\subsubsection{Adaptive ensembles}
\label{sec:l:ensembles}
\index{adaptive ensembles}

The most popular evolving technique for handling concept drift is \emph{classifier ensemble}. 
Classification outputs of several models are combined  or selected to get a final decision. 
The combination or selection rules are often called fusion rules. 

There is a number of ensembles for concept drift, where the ideas are not specific to particular type of base learners (although some studies are limited to testing one base learner) \cite{Kolter07,Stanley03,Street01,Wang03,Tsymbal08,Karnick08,Scholz07,Becker07,Nishida05,Wang06,Fan04,Zhang08,Rodriguez08}. 
There are also base learner specific ensembles. In those classifier combination rules usually depend on the base learner specific parameters of the learned models: \cite{Klinkenberg00,Klinkenberg04} with SVM, \cite{Wu05} with Gaussian mixture models, \cite{Raudys07} with perceptrons, \cite{Law05} with kNN. 

In both cases \emph{adaptivity} is achieved by fusion rules, i.e. how the weights are assigned to the individual model outputs at each point in time. 
In a discrete case an output of a single model might be selected.
In this case all except one models get zero weights. 
The weight indicates the `competence' of a base learner, expected in the `nearest future' (future assumption A.1). 
The weight is usually a function of the historical performance \cite{Kolter07,Stanley03,Street01,Wang03,Karnick08,Becker07,Nishida05,Zhang08,Rodriguez08} in the past or estimated performance using selective cross validation \cite{Tsymbal08,Scholz07,Wang06,Fan04,Law05} or base learner specific performance estimates \cite{Klinkenberg00,Klinkenberg04,Wu05,Raudys07}. 
The historical evaluation is restricted to sudden and incremental drifts, while cross validation allow taking into account gradual drifts and reoccurring contexts. 

In adaptive ensemble learners much attention is drawn to model evaluation and fusion rules (A.4), while little attention is drawn to the model construction (A.3). Still there is a number of options how to build diverse base classifiers. Usually the implicit aim is to have at least one classifier in the ensemble trained for each distinct concept. 
This can be achieved using different training set selection strategies.

The straightforward approach is to divide historical data into blocks, which include instances sequential in time. Often these blocks are non overlapping \cite{Street01,Wang03,Tsymbal08,Becker07,Nishida05,Karnick08,Klinkenberg00,Klinkenberg04,Law05}, sometimes overlapping \cite{Fan04}. These techniques are suitable for sudden and to some extent to incremental drifts, they favor reoccurring contexts. Another approach is using different sized training windows \cite{Kolter07,Stanley03,Scholz07,Raudys07}, which implicitly assume that once off sudden drift has happened. Training windows are overlapping sequential blocks of instances, but all of them have fixed ending `now' (time $t$). 
The individual models in ensembles can also be constructed using non sequential instance selection \cite{Wang06}. 
This technique is more suitable to gradual drift, as well as reoccurring contexts. 


Another approach to building diverse base classifiers is to use the same training data, but different types of base learners (e.g. SVM, decision tree, Naive Bayes) \cite{Zhang08,Rodriguez08}. 

All these techniques build individual models from what has already been seen in the past. 
In principle base classifiers can also be built adding unseen data, for instance noise or unlabeled testing data, which is listed as our future work.

\subsubsection{Instance weighting}

Instance weighting weighting methods make another group of evolving adaptation techniques. The algorithms can consist of a single learner \cite{Koychev00,Zhang08,Oommen06} or an ensemble \cite{Gao08,Bifet09,Chu04}, but the adaptivity here is achieved not by combination rules, but by systematic training set formation. Ideas from boosting \cite{Freund99} 
are often employed, giving more attention to the instances which were misclassified. 

\subsubsection{Feature space}

There are models, manipulating feature space to achieve adaptivity. \cite{Forman06} uses ideas from transfer learning to achieve adaptivity. New features are added to the training instances, which contain information from the past model performances. \cite{Black02} augments the feature space by a time stamp. \cite{Katakis06,Wenerstrom06} use dynamic feature space over time. 
In \cite{Anagnostopoulos08} the variables to observe next are adaptively selected.

\subsubsection{Base model specific}

There are also models to be mentioned, where adaptivity is achieved by managing specific model parameters or design. \cite{Nunez07} maintain variable training window via adjusting internal structure of decision trees. Regression parameters are being adjusted in \cite{Kelly99}. Past support vectors are transfered and combined with the recent training data in \cite{Syed99}. The later examples illustrate the variety of possible specific model designs. 

\subsection{Learners with triggers}

Another group of methods uses triggers, which determine how the models or sampling should be changed at a given time. 

\subsubsection{Change detectors}
\index{change detectors}

The most popular trigger technique is change detection, which is often implicitly related to a sudden drift. Change detection can be based on monitoring the raw data \cite{Bifet07,Patist07}, the parameters of the learners \cite{Su08} 
or the outputs (error) of the learners \cite{Gama04,Garcia06,Nishida07}. \cite{Dries09} develop change detection methods in each of the three categories. The detection methods usually cut the training window at change point, although the change point and training window might not be the same \cite{Zliobaite08semofor}.

\subsubsection{Training windows}
\index{training windows}

There are methods using heuristics for determining training window size \cite{Widmer96,Lazarescu04,Bach08,Yang06}. The heuristics is related to error monitoring. The training window is determined using look up table principles, where there is an action for each possible value of a trigger. 
There also are base learning specific methods, for determining training windows \cite{Hulten01,Zhang08,Last02,Tsai09}. The window size is also determined based on historical accuracy.

\subsubsection{Adaptive sampling}
\index{adaptive sampling}

The listed trigger based methods were using training windows. Another group of trigger based methods use instance selection. The incoming testing instances (unlabeled) are inspected. Based on the relation between the testing instance and predefined prototypes \cite{Delany05,Katakis09,Klinkenberg05,Yang08} or historical training instances directly\cite{Pourkashani08,Hashemi07,Lazarescu03,Beringer07} a training set for a given instance is selected.  








\subsection{Discussion}

In Table \ref{tab:algsummary} we provide a summary of the listed algorithms. 
The properties are structured according to the four design assumptions, which were discussed in Section \ref{sec:designassumptions}. The categorization is based on our interpretation of the methods. 

Change detectors and ensembles are the two most popular techniques. Change detectors are naturally suitable for the data where sudden drift is expected. Ensembles, on the other hand, are more flexible in terms of change type, while they can be slower in reaction in case of a sudden drift. 

We overviewed general methods for handling concept drift in supervised learning. A discussion of specific applications will follow in Section \ref{sec:applications}. Before proceeding to applications let us look at the broader context of learning with changing data.

\begin{sidewaystable}
\centering
\caption{Summary of concept drift responsive algorithms}
\begin{tabular}{|c|c|c|c|c|c|}
\hline
 							& 							& \multicolumn{ 4}{c|}{Design Assumptions} \\ \hline
papers				& trigger 			& A.1 Future	& A.2 Change & A.3 Learner & A.4 Selection \\ \hline
\cite{Kolter07,Stanley03} 
& 						& 				& 					& tr. windows & 	 \\
\cite{Street01,Wang03,Karnick08,Becker07,Nishida05}
& no (ensemb.)& last 		& sud./ inc.& time blocks & hist. err \\ 
\cite{Zhang08,Rodriguez08}
& 						& 				&						& learner sp. & 	\\ \hline
\cite{Scholz07,Fan04,Law05}
&							& last 		& sud./ inc. & tr. windows & 	 \\ 
\cite{Tsymbal08}
& no (ensemb.)& estim. 	& grad./ sud.& time blocks & cross v. \\ 
\cite{Wang06}
& 						& estim. 	& grad./ sud.& inst. select.& 	\\ \hline
\cite{Klinkenberg00,Klinkenberg04}
& 						& last 		& sud./ inc. & time blocks & 	 \\ 
\cite{Raudys07}
& no (ensemb.)& last 		& sud./ inc. & tr. windows & learner dep. \\ 
\cite{Wu05}
& 						& pred. 	& grad./ sud.& learner sp. & 		\\ \hline
\cite{Koychev00,Zhang08,Oommen06}
& no (inst. wght.)& last 	& inc. 				& inst. wght. & hist. err \\ 
\cite{Gao08,Bifet09,Chu04}
& 								& 			& 						& 						& 	\\ \hline
\cite{Forman06,Katakis06,Black02,Wenerstrom06}
& no (feature sp.)& last 	& various 		& feature sp. & hist. err \\ \hline
\cite{Kelly99,Syed99}
& no 							& last 	& various 		& learner sp. & hist. err \\ 
\cite{Nunez07}
& 								& 			& 						& 						& learner dep. \\ \hline \hline
\cite{Bifet07,Patist07,Gama04,Garcia06,Nishida07,Dries09}
& yes (ch. detec.)& last 	& sud. 				& tr. windows & hist. err \\ \hline
\cite{Widmer96,Lazarescu04,Bach08,Yang06}
& yes (windows) 	& last 	& sud. /inc. 	& tr. windows & hist. err \\ 
\cite{Hulten01,Zhang08,Last02,Tsai09}
& 								& 			& 						& learner sp. & learner dep. \\ \hline
\cite{Delany05,Katakis09,Klinkenberg05,Yang08}
& yes (inst. sel.)& estim.& grad./ sud. & inst. sel. 	& prototyping \\ 
\cite{Pourkashani08,Hashemi07,Lazarescu03,Beringer07}
& 								& 			& 						& 						& cross v. \\ \hline
\end{tabular}
\label{tab:algsummary}
\end{sidewaystable}

\section{Related Research Areas}
\label{sec:areas}


After reviewing the adaptive techniques for supervised learning, which were mostly developed in data mining and machine learning communities, we now give an interdisciplinary perspective of the concept drift problem. 
In this section we point the `neighboring' research fields. 
We pick the works, which are not necessary the `key' references in these fields, but the ones which touch a problem of dataset change. 

\begin{figure}
\centering
\includegraphics[width=0.7\linewidth]{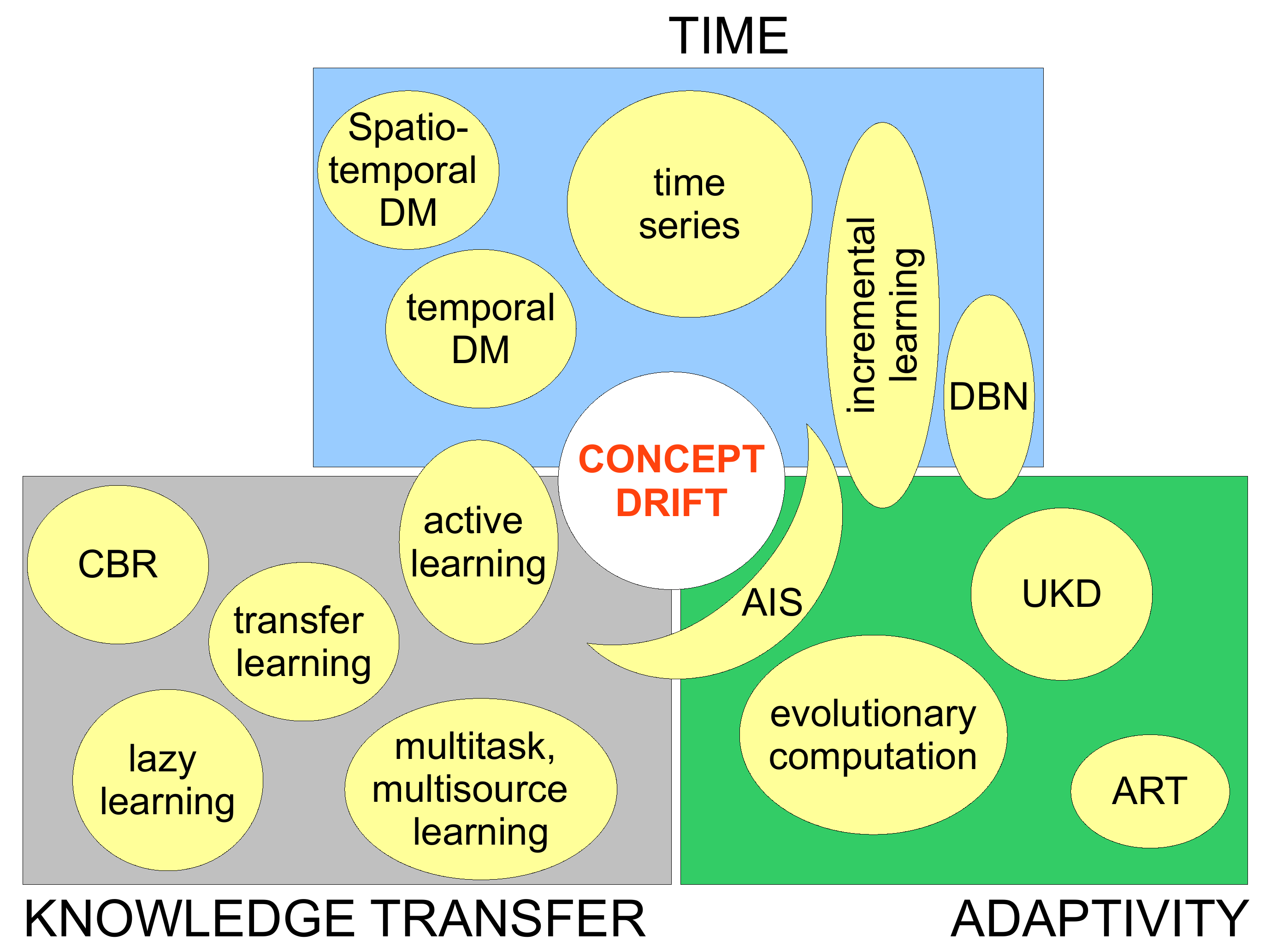}
\caption{Categorization of the related areas. AIS - Artificial Immune Systems; DBN - Dynamic Bayesian Networks; UKD - Ubiquitous Knowledge Discovery; ART - Adaptive Resonance Theory; CBR - Case Based Reasoning.}
\label{fig:relatedareas}	
\end{figure}

We present the research fields in three categories, which we identified as connections with the concept drift problem: time, knowledge transfer and adaptivity. In Figure \ref{fig:relatedareas} we position the related areas within these three categories. We discuss them in the following sections. 

\subsection{Time context}

Time context in concept drifting problems means that the data is sequential in time and the models are also associated with time and need to be continuously updated. 
There are research fields focusing on the aspects of model update primarily for a stationary data.

\emph{Incremental learning} focuses on machine learning the tasks, where all the training data is not available at once \cite{Fisher88,Carrier00}. The data is received over time thus the models need to be updated or retrained, to increase the accuracy. Schlimmer and Granger \cite{Schlimmer86} introduced the assumption of concept change in incremental learning context.
\index{incremental learning}

Over decades \emph{incremental learning} area became less active. 
It was gradually overtaken by \emph{data stream mining}, where the data flow is continuous an rapid \cite{Gaber05}. 
Data stream mining focuses on the processing speed and complexity,
thus naturally the attention toward timely \emph{change detection} \cite{Lu04} including \emph{anomaly detection} \cite{Chandola09} has increased.

\emph{Spatio - temporal data mining} deals with database models to accommodate temporal aspects \cite{Pelekis04,Roddick02}. \emph{Temporal data mining} \cite{Laxman06} incorporates time dimension into data mining process.
\index{temporal data mining}
\index{spatio-temporal data mining}

\emph{Dynamic Bayesian networks} are causal models assuming forward relation between the variables in time \cite{Cho08}. 
\index{dynamic bayesian networks}

Finally, in \emph{time series analysis} non stationarity is handled using ARIMA models \cite{Box90}.\\
\index{time series analysis}


\subsection{Knowledge transfer}

Knowledge transfer means that regularly there is a potential difference between the distribution of training data and the data to which the models will be applied (testing data). Thus the information from the old data needs to be adapted to fit to the new data. In concept drift problem this discrepancy arises in time, due to changes in the data generating process. However, a dataset shift can have a number of other reasons, for instance, sample selection bias \cite{Bickel07}
, domain shift due to changes in measurements, model shift due to imbalance of data \cite{Candela09}, discrimination in decision making \cite{Kamiran09}, which are out of the scope of this thesis. In addition, the knowledge from related problem might be transfered to solve a related one.
\index{dataset shift}

\emph{Case based reasoning} (CBR) \cite{Lopez97} is the process of solving new problems based on the solutions of similar past problems. Generally CBR can be treated as \emph{lazy learning}. Lazy learning does not build generalizing models, but maintain a database of reference data and uses the relevant past data only when a related query is made \cite{Aha97}. In this domain Aha \cite{Aha91} introduced noise tolerant instance based algorithms, IB3 was the first instance based technique capable of handling concept drift.
\index{case based reasoning}
\index{lazy learning}

A great part of lazy learning research is devoted to \emph{instance selection} methods to increase accuracy. There is another related instance selection research area (not necessarily lazy learning) aiming to reduce the learning complexity by data reduction \cite{Reinartz02}.

In machine learning the process of applying the knowledge gained on solving a similar problem is referred as \emph{transfer learning} \cite{Rosenstein05} or inductive transfer. The ideas of inductive transfer were extended to temporal representation and used for learning under concept drift \cite{Forman06}.
\index{transfer learning}

Adaptive knowledge transfer has been exploited in \emph{multitask learning} \cite{Merrick09} and \emph{learning from multiple sources} \cite{Crammer08,Mansour08}. Non stationarity problem in machine learning community is sometimes called \emph{covariate shift} \cite{Blitzer07,Jiang07}.
\index{covariate shift}
\index{multitask learning}
\index{learning from multiple sources}

Finally, a field of \emph{active learning} \cite{Settles09} is remotely related to the problem of concept drift. In active learning the data is labeled on demand, the methods select the instances which need to be labeled to make the learner more accurate or reduce labeling costs. The relation to concept drift problem is in the ways the methods identify, how well the unlabeled instances correspond to a particular concept.
\index{active learning}


\subsection{Model adaptivity}

Model adaptivity here means the models which have the properties of adaptation incorporated into learning. The adaptation might be to a change, as in concept drift problem. Adaptation can also mean the learning process (in stationary or non stationary environment), when the accuracy of the model is increasing along with more incoming examples. 

\emph{Artificial immune systems} (AIS) are inspired by immunology \cite{Freitas07}. They are adaptive to changes like biological immune systems. AIS use evolutionary computation and memory to learn to recognize changing patterns.
\index{artificial immune systems}

\emph{Adaptive resonance theory}, dating back 30 years \cite{Grossberg76}, is based on the model of information processing by the brain \cite{Carpenter02}. Having self-adjusting memory as one of the desired system properties.
\index{adaptive resonance theory}

In \emph{evolutionary computation} dynamic optimization problems are actively studied \cite{Morrison04}. The goal is track the optima which is dynamically changing in time. The major approaches are related to maintaining and enhancing diversity, expecting that once the optima changes, there are suitable models available withing the pool \cite{Yang08evo}. A a next step in this direction is to seek for a relation between the past models and current task \cite{Richter09}. In \cite{Rohlfshagen09} a relation between the change type and magnitude and the evolutionary algorithm is introduced.
\index{evolutionary computation}

\emph{Ubiquitous knowledge discovery} is an emerging area, which focuses on learning in distributed and mobile systems \cite{May08}. The systems work in environment, they need to be intelligent and adaptive. The objects of UKD systems exist in time and space in a dynamically changing environment, they can change location and might appear or disappear. The objects have information processing capabilities, know only their local spatio-temporal environment, act under real-time constraints and are able to exchange information with other objects. These objects are humans, animals, and, increasingly, computing devices.\\
\index{ubiquitous knowledge discovery}

To sum, the problem of change is far not limited to data mining and machine learning community. 
Concept drift problem lies in all three dimensions: time dimension, need for adaptivity and knowledge transfer. 

\section{Applications}
\label{sec:applications}

In this section we survey applications, where concept drift problem is relevant in both supervised and unsupervised learning. 
We present the real life problem, discuss the sources of a drift and the actual learning tasks in the context of these problems. 

We find four general types of applications: monitoring control, personal assistance, decision making and artificial intelligence. \emph{Monitoring and control} often employs unsupervised learning, which detects abnormal behavior. It includes detection of adversary activities on the web, computer networks, telecommunications, financial transactions. \emph{Personal assistance and information} applications include recommender systems, categorization and organization of textual information, customer profiling for marketing. \emph{Decision making} includes diagnostics, evaluation of creditworthiness. The `ground truth' is usually delayed, i.e. the true answer whether the decision was correct becomes available only after certain time. \emph{Artificial intelligence} applications include a wide spectrum of moving and stationary systems, which interact with changing environment, for instance robots, mobile vehicles, smart household appliances.

We define five dimensions, relevant to the applications facing concept drift:
\begin{enumerate}
\item the speed of learning and output,
\item classification or prediction accuracy,
\item costs of mistakes,
\item true labels,
\item adversary activities. 
\end{enumerate}
The speed of learning output means what is a relative volume of data and how fast the decision needs to be made. For example, in credit card fraud detection the decision needs to be fast to stop the crime and the data loads are huge, while in credit evaluation a decision regarding the credit can be made even in a few days time. In both cases adversary activities to cheat the system might be expected, while adversary activities in diagnostics would make less sense. The precise accuracy in diagnostics is generally much more significant than in movie recommendations, moreover, in movie recommendations the decision might be `soft' in a sense the viewer is not always deterministic, which movie he or she liked more. 

Our global interpretation of the four types of applications in accord with these dimensions is provided in Table \ref{tab:applicationdimensions}.

\begin{table}[th]
\centering
\caption{Types of applications with concept drift.}
{\small
\begin{tabularx}{\textwidth}{|X|lllll|}
\hline
	&Decision	&Accuracy		&Costs	of	&Labels		&Adversary\\			
	&speed		&						& mistakes	&					&					\\
\hline
1. Monitoring $\&$ control	&high	&approximate &medium	&hard	&active \\
2. Assistance $\&$ information &medium	&approximate &low	&soft	&low \\
3. Decision making &low &precise &high &delayed &possible	\\
4. AI and robotics &high &precise &high &hard &low \\
\hline
\end{tabularx}
}
\label{tab:applicationdimensions}
\end{table}

In the following sections we discuss each of the application types separately and give arguments for the choices we made in the table. 



\subsection{Monitoring and Control}

In monitoring and control applications the data volumes are large and it needs to be processed in real time. Two types of tasks can be distinguished: prevention and protection against adversary actions, and monitoring for management purposes. 

\subsubsection{Monitoring against adversary actions}

Monitoring against adversary actions is often an unsupervised learning task or one class classification, where the properties of `normal behavior' are well defined, while the properties of attacks can differ and change from case to case. Classes are typically highly imbalanced with a few real attacks.

\paragraph{Computer security.} \emph{Intrusion detection} is one of the typical monitoring problems. That is a detection of unwanted access to computer systems mainly through network (e.g. internet). There are passive intrusion detection systems, which only detect and alert the owner, and active systems, which take protective action. In both cases here we refer only to a detection part. 

Adversary actions is the primary source of concept drift in intrusion detection. The attackers try to invent new ways how to attack, which would overcome the existing security. The secondary source of concept drift is technological progress in time, when more advanced and powerful machines are created, they become accessible to intruders. `Normal' behavior can also change over time. 

Lane and Brodley \cite{Lane99} explicitly formulated the problem of concept drift in intrusion detection a decade ago. They presented a detection system using instance based learning. Current research directions and problematics in intrusion detection can be found in a general review \cite{Patcha07}. From supervised learning, lately, ensemble techniques have been proposed \cite{Masud09}. Artificial immune systems are widely considered for intrusion detection\cite{Kim07}. 

\paragraph{Telecommunications.} Adversary behavior also applies to \emph{telecommunications industry}, both intrusion and fraud. 
Mobile masquerade detection problem \cite{Mazhelis07} from research perspective is closely related to intrusion detection. The goal is to prevent adversaries from unauthorized access to a private data. The sources of concept drift are again twofold: adversary behavior trying to overcome the control as well as changing behavior of legitimate users. 
Fraud detection and prevention in telecommunication industries  \cite{Hilas09} is also subject to concept drift due to similar reasons. 

\paragraph{Finance.} In \emph{financial sector} data mining techniques are employed to monitor streams of financial transactions (credit cards, internet banking) to alert for possible frauds. 
Insider trading in stock market is one more application. 

Both supervised and unsupervised learning techniques are used \cite{Bolton02} for detection of fraudulent transactions. The data labeling might be imprecise due to unnoticed frauds, legitimate transactions might be misinterpreted and the imbalance of the classes is very high (few frauds as compared to legitimate actions). Concept drift in user behavior is one of the challenges. 

Insider trading is trading in stock market based on non-public information about the company, in most countries it is prohibited by law. Inside information can come in many forms: knowledge of a corporate takeover, a terrorist attack, unexpectedly poor earnings, the FDA's acceptance of a new drug \cite{Donoho04}, inside trading disadvantages regular investors. There is a potential for concept drift, since the inside traders would try to come up with novel ways to distribute the transactions in order to hide.


\subsubsection{Monitoring for management}

Monitoring for management usually uses streaming data from sensors. It is also characterized by high volumes of data and real time decision making; however, adversary cases usually are not present. 

\paragraph{Transportation.} \emph{Traffic management} systems use data mining to determine traffic states \cite{Crespo05}, e.g. car density in a particular area, accidents. Traffic control centers are the end users of such systems. Transportation systems are dynamic (always moving).The traffic patterns are changing seasonally as well as permanently, thus the systems have to be able to handle concept drift.  

Data mining can also be employed for prediction of public transportation travel time \cite{Moreira08}, which is relevant for scheduling and planning. The task is also subject to concept drift due to traffic patterns, human driver factors, irregular seasonality. 

\paragraph{Positioning.} Concept drift is also relevant in remote sensing in \emph{fixed geographic locations}. Interactive road tracking is an image understanding system to assist a cartographer annotating road segments in aerial photographs \cite{Zhou08}. In this problem change detection comes into play when generalizing to different roads over time. 
In place recognition \cite{Luo07} or activity recognition \cite{Liao07} dynamics of the environment cause concept drift in the learned models. 

Climate patterns, such as floods, are expected to be stationary, but the detection systems have to incorporate not regular reoccurring contexts. In a light of a climate change the systems might benefit from adaptive techniques, for instance, sliding window training \cite{Mahmud09}. In \cite{Krause07} the authors use active learning of non stationary Gaussian process for river monitoring.

\paragraph{Industrial monitoring.}
In \emph{production monitoring} human factor can be the source of concept drift. Consider a boiler used for heat production. The fuel feeding and burning stages might depend on individual habits of a boiler operator, when the fuel is manually input into the system \cite{Bakker09}. The control task is to identify the start and end of the fuel feeding, thus algorithms should be equipped with mechanisms to handle concept drift. 

In \emph{service monitoring} changing behavior of the users can be the source of a drift. For example, data mining is used to detect accidents or defects in telecommunication network \cite{Pawling07}. A change in call volumes may be the results of an increased number of people trying to call friends or family to tell them what is happening or a decrease in network
usage caused by people being unable to use the network. Or the change might be unrelated to the telecommunication network at all. The fault detection techniques have to be able to handle such anomalies.

\subsection{Personal Assistance and Information}

These applications mainly organize and/or personalize the flow of information. The applications can be categorized into individual assistance for personal use, customer profiling for business (marketing) and public or specified information. In any case, the class labels are mostly `soft' and the costs of mistake are relatively low. For example, if a movie recommendation is wrong it's not a world disaster and even the user himself or herself might not know for sure, which of the two given movies he or she likes more.

\subsubsection{Personal assistance}

Personal assistance applications deal with user modeling aiming to personalize the flow of information, which is referred as \emph{information filtering}. A rich technical presentation on user modeling can be found in \cite{Gauch07}. One of the primary applications of user modeling is representation of queries, news, blog entries with respect to current user interests.
Changes in user interests over time are the main cause of concept drift. 
\index{information filtering}
\index{user modeling}

Large part of personal assistance applications are related to \emph{textual data}. 
The problem of concept drift has been addressed in news story classification \cite{Widyantoro05,Billsus99} or document categorization \cite{Lebanon08,Klinkenberg98,Mourao08}. 
\cite{Katakis08} in a light of changing user interests address the issue of reoccurring contexts. 
Drifting user interests are relevant in building personal assistance in digital libraries \cite{Hasan08} or networked media organizer \cite{Flasch07}. 

There is also a large body of research addressing web personalization and dynamics \cite{Yamaguchi99,Scanlan08,Silva07,deBra03}, which is again subject to drifting user interests. In contrast to end user text mining discussed before, here mostly interim system data (logs) is mined. 

Finally, concept drift problem is highly relevant for spam filtering \cite{Delany06,Riverola07}. First of all there are adversary actions (spamming) in contrast to the personal assistance applications listed before. That means the senders are actively trying to overcome the filters therefore the content changes rapidly. Adversaries are intelligent and adaptive.
Spam types are subject to seasonality and popularity of the topics or merchandises. There is a drift in the amount of spam over time, as well as in the content of the classes \cite{Fawcett03}. Spam messages are disjunctive in content. Besides, personal interpretation of what is spam might differ and change. 

\subsubsection{Customer profiling}

For customer profiling aggregated data from many users is mined. The goal is to segment the customers according to their interests. Since individual interests are changing over time, customer profiling algorithms should take this non stationarity into account. 

Direct marketing is one of the applications. Adaptive data mining methods are used in customer segmentation based on product (cars) preferences \cite{Crespo05} or service use (telecommunications) \cite{Black02}. Lately in addition to similarity measures between individual customers social network analysis has been employed into customer segmentation \cite{Lathia08}. It is observed that user interests do not evolve simultaneously. The users that used to have similar interests in the past might no longer share the interests in the future. The authors model this as an evolving graph. 
Adaptivity is also relevant to association rule mining applied to shopping basket identification and analysis \cite{Rozsypal05}. 

Automatic recommendations can be related to both customer profiling and personal assistance. 
The recommender systems are characterized by sparsity of data. For example, there are only a few movie ratings per user, while the recommendations need to be inferred over the while movie pool. 
The publicity of recommender systems research has increased rapidly with a NetFlix movie recommendation competition. The winners used temporal aspect as one of the keys to the problem \cite{Koren09,Bellkor08}. Three sources of drift were noted movie biases (popularity changes over time), user bias (natural drift of users' rating scale benchmarking to the recent ratings) and changes in user preferences. There are earlier works on recommender systems in which changes over time were addressed \cite{Ding05} via time weighting.


\subsubsection{Information}

Information applications are related to changes in data distribution over time, which is sometimes referred as virtual drift in concept drift literature \cite{Widmer93}. Then changes in class assignment is called real drift. Virtual drift would typically occur over longer period of time. For example, in news recommendation system, the news about meat prices in New Zealand suddenly become relevant for Kate (the label changes, but the document comes from the same distribution as before). It might happen that the consumers in New Zealand would switch from pork to beef, thus the distribution of articles about meat would change independently from Kate's interests.

\paragraph{Document organization} is the first category of information applications. Given e-mail, news or document streams, the task is to extract meaningful structures, organize the data into topics. Temporal order is necessary for making sense. The topics themselves and even the vocabulary for particular topics change in time. 

The state of the art Latent Dirichlet Allocation model for probabilistic document corpus modeling was recently equipped with a time dimension \cite{Blei06,Wang08}.
In \cite{Blei06} the dynamics of scientific topics articles of Science magazine from 1881 to 1999 (120 years) was analyzed, the emergence, peak and decline of topics was showed, the topic vocabulary representation was build. \cite{Yang06} incorporated the time stamp into the static model.
\cite{Kleinberg02} presented a method for organization of e-mail messages, to provide a framework for content analysis.   Intuitively this is similar to including time feature into the original observation. 

\paragraph{Economics.} Concept drift is relevant in making macroeconomic forecasts \cite{Giacomini06}, predicting the phases of a business cycle \cite{Klinkenberg05}. The data is drifting primary due to large number of influencing factors, which are not feasible to be taken into prediction models. Due to the same reason financial time series are known to be non stationary to predict  \cite{Harries95}. 

In \textbf{business management}, in particular, software project management, careful planning can be inaccurate if concept drift is not taken into account. \cite{Ekanayake09} employ data mining models for project time prediction, the models are equipped with concept drift handling techniques.

\subsection{Decision Making} 

Decision making and diagnostics applications usually involve limited amount of data (might be sequential or time stamped). 
Decisions are not required to be made in real time, thus the applied models might be computationally expensive. But high accuracy is essential in these applications and the costs of mistakes are large. 

\paragraph{Finance.}
\emph{Bankruptcy prediction} or individual credit scoring is typically considered to be a stationary problem \cite{Kumar07}. However, in these problems concept drift is closely related to a hidden context \cite{Harries98}, changes in context, which is not observed or measured in the original model. The need for different models for bankruptcy prediction under different economic conditions was acknowledged and proposed in \cite{Sung99}. The need for models to be able to deal with non stationarity has been rarely acknowledged \cite{Horta09}. Although concept drift problem is present, adversaries might make use of full adaptivity of the models. Thus offline adaptivity, which would be restricted to already seen subtypes of customers, is needed \cite{Zliobaite09credit}.  

\paragraph{Biomedical applications} can be subject to concept drift due to adaptive nature of microorganisms \cite{Song08,Tsymbal08}.
The effect of antibiotics to a patient is often naturally diminishing over time, since microorganisms mutate and evolutionary develop antibiotic resistance. If a patient is treated with antibiotic when it is not neccesary, a resistance might develop and antibiotics might no longer help when they are really needed.

Clinical studies and systems need adaptivity mechanisms to changes caused by human demographics \cite{Kukar03,Gago07}. The changes in disease progression can also be triggered by changes in a drug being used \cite{Black04}.
In incremental drug discovery experiments the drift between training and testing sets can caused by non uniform sampling \cite{Forman02}. 

Data mining can be used to discover emerging resistance and monitor nonsomnical infections in hospitals (the infections which result from the treatment) \cite{Jermaine08}. Given patient and microbiology data as an input, the task is to model the resistance. The resistance changes over time. 

Finally, concept drift occurs in biometric authentication \cite{Yampolskiy07,Poh09}. The drift can be caused by changing physiological factors, for example growing beard. Like in credit applications, here adaptivity of the algorithms should be used with caution, due to potential adversary behavior.

\subsection{AI and Robotics}

In AI applications the problem of concept drift is often called dynamic environment. The objects learn how to interact with the environment and since the environment is changing, the learners need to be adaptive.

\subsubsection{Mobile systems and robotics}

\emph{Ubiquitous Knowledge Discovery} (UKD) deals with the distributed and mobile systems, operating in a complex, dynamic and unstable environment. The word 'ubiquitous' means distributed at a time. Navigation systems, vehicle monitoring, household management systems, music mining are examples of UKD.

DARPA navigation challenge was presented in \cite{Thrun06}. A winning entry in 2005 used online learning for road image classification into drivable and not drivable. They used an adaptive Mixture of Gaussians, for gradual adaptation they were adjusting the internal Gaussian and rapid adaptation by replacement of the Gaussians with the new ones. The needed speed of adaptation would depend on the road conditions. 

Adaptivity to changing environment has been addressed in robotics \cite{Procopio09}, for instance in designing a player for robot soccer \cite{Lattner06}.
\index{robotics}

\subsubsection{Intelligent systems}

`Smart' home systems \cite{Rashidi09} or intelligent household appliances \cite{Anguita01} also need to be adaptive to changing environment and user needs.

\subsubsection{Virtual reality}

Finally, virtual reality needs mechanisms to take concept drift into account. In computer game design \cite{Charles05} adversary actions of the players (cheating) might be one of the drift sources. 
In flight simulation the strategies and skills differ across different users \cite{Harries98}.

In Table \ref{tab:applications} we summarize the discussed applications with concept drift.

\begin{sidewaystable}
\caption{Summary of applications with concept drift}
\begin{tabular}{|c|c|c|c|c|}
\hline
\multicolumn{3}{|c|}{CATEGORIES} 									& APPLICATIONS 									& REFERENCES \\ 
\hline
						& against 		&	computer security			&	intrusion detection						&	\cite{Lane99,Masud09,Kim07}			\\
													\cline{3-5}
Monitoring 	&	adversaries	&	telecommunications		&	intrusion detection, fraud		&	\cite{Mazhelis07,Hilas09} \\
													\cline{3-5}
and					&							&	finance								& fraud, insider trading				& \cite{Bolton02,Donoho04}\\
						\cline{2-5}
control			& for					& transportation				& traffic management							& \cite{Crespo05,Moreira08}\\
													\cline{3-5}
						& management 	&	positioning						&	place, activity recognition		& \cite{Zhou08,Luo07,Liao07}\\
													\cline{3-5}
						& 						&	industrial mon.				&	boiler control, telecom mon.	& \cite{Bakker09,Pawling07}\\
\hline						
						& 						& textual information		& news, document classification & \cite{Widyantoro05,Billsus99,Lebanon08,Klinkenberg98,Mourao08} \\
						&	personal		&												& spam categorization						& \cite{Delany06,Riverola07} \\
													\cline{3-5}
						&	assistance	& web										& web personalization						& \cite{Yamaguchi99,Scanlan08,Silva07,deBra03} \\									
Assistance	&							& 											& libraries, media							& \cite{Hasan08,Flasch07} \\
						\cline{2-5}
and					& customer		& marketing							& customer segmentation					& \cite{Crespo05,Black02,Lathia08,Rozsypal05} \\
													\cline{3-5}
information	& profiling		& recommender systems		&	movie recommendations					& \cite{Koren09,Bellkor08,Ding05} \\
						\cline{2-5}								
						& 						& document organization	& articles, mail								&	\cite{Blei06,Wang08,Yang06,Kleinberg02} \\			
													\cline{3-5}
						& information	& economics 						& macroeconomics, forecasting		& \cite{Giacomini06,Klinkenberg05,Harries95} \\
													\cline{3-5}
						& 						& project management		& software project mgmt.				& \cite{Ekanayake09} \\
\hline
						& finance			&	creditworthiness			& bankruptcy prediction					& \cite{Sung99,Horta09,Zliobaite09credit} \\
						\cline{2-5}
Decision		& biomedicine	& drug research					& antibiotic res., drug disc. 	& \cite{Tsymbal08,Forman02,Jermaine08} \\
													\cline{3-5}
making 			&							& clinical research			&	disease monitoring						&	\cite{Kukar03,Gago07,Black04} \\
						\cline{2-5}
						& security		& authentication				& biometrics										& \cite{Yampolskiy07,Poh09} \\
\hline
AI					& 						& mobile systems				& robots, vehicles							& \cite{Thrun06,Procopio09,Lattner06} \\
													\cline{3-5}
and					& 						& intelligent systems		& `smart' home, appliances			& \cite{Rashidi09,Anguita01} \\
													\cline{3-5}
robotics		&							& virtual reality				& computer games, flight sim. 	& \cite{Charles05,Harries98} \\
\hline
\end{tabular}
\label{tab:applications}
\end{sidewaystable}

\section{Terminology}
\label{sec:terminology}
\index{terminology}

Concept drift is relatively new research field and the terminology is not yet fixed. 
Moreover, the problem of shifting data is discovered and handled in very broad domain area. 
With the loads of data more and more attention is drawn to the differences between training and testing data distributions. 
We provide alternative terminology in Table \ref{tab:terminology}.

\begin{table}[h]
\centering
\caption{Terminology across research fields.}
\begin{tabular}{ll}
\hline
Data mining 							& concept drift \\
Machine learning 					& concept drift, covariate shift \\
Evolutionary computation 	& changing environment \\
AI and Robotics						& dynamic environment \\
Statistics, time series 	& non stationarity \\
Databases									& concept drift, load shedding \\
Information retrieval 		& temporal evolution \\ 
\hline
\end{tabular}
\label{tab:terminology}
\end{table}




\section{Concluding Remarks}

We provided an overview of the available concept drift responsive techniques and real learning tasks, where concept drift problem is relevant. 

The problem of concept drift is very broad. 
There has been plenty of general research and attempts to understand the phenomena. 
Generalization is not possible without assumptions about the nature of the change. 
This depends on the data and the problem. 
The focus on applications has been limited so far. Lack of real data?

We argue that the challenges are different for different types of applications, see Table \ref{tab:applicationdimensions}. 
How quickly does the data change? Is it worth complicating the model? It is if we deal with 100 years history of Science papers. 
Do we want full model adaptability? Is it secure? 
Wouldn't a simple training window be enough in the most practical cases? 
Maybe focusing on selecting a proper the base model is essential? 

In our opinion, focus on specific models for specific problems is prospective.

\bibliographystyle{plain}
\bibliography{thesis_bibliography}

\end{document}